\documentclass[conference]{IEEEtran}
\usepackage{cite}
\usepackage{amsmath,amssymb,amsfonts}
\usepackage{algorithmic}
\usepackage{graphicx}
\usepackage{textcomp}
\usepackage{xcolor}
\usepackage{subcaption}
\usepackage[font=small,skip=3pt]{caption}
\usepackage{multirow}
\def\BibTeX{{\rm B\kern-.05em{\sc i\kern-.025em b}\kern-.08em
		T\kern-.1667em\lower.7ex\hbox{E}\kern-.125emX}}

\def\ps@IEEEtitlepagestyle{%
	\def\@oddfoot{\mycopyrightnotice}%
	\def\@evenfoot{}%
}
\def\mycopyrightnotice{%
	{\footnotesize \textcopyright 2019 IEEE. Personal use of this material is permitted. Permission from IEEE must be obtained for all other uses, in any current or future media, including reprinting/republishing this material for advertising or promotional purposes, creating new collective works, for resale or redistribution to servers or lists, or reuse of any copyrighted component of this work in other works. DOI: 10.1109/IJCNN.2019.8852414 \hfill}% <--- Change here
	\gdef\mycopyrightnotice{}% just in case
}

\begin{document}

%	20xx IEEE. Personal use of this material is permitted. Permission from IEEE must be obtained for all other uses, in any current or future media, including reprinting/republishing this material for advertising or promotional purposes, creating new collective works, for resale or redistribution to servers or lists, or reuse of any copyrighted component of this work in other works.
%	\IEEEoverridecommandlockouts
%	\IEEEpubid{\makebox[\columnwidth]{xxx-xxx-xxx-xxx\$31.00 \copyright 2019 IEEE \hfill }
%		\hspace{\columnsep}\makebox[\columnwidth]{\hfill }}
	
%	\IEEEoverridecommandlockouts
%	\IEEEpubid{\makebox[\columnwidth]{ \copyright 2019 IEEE. Personal use of this material is permitted. Permission from IEEE must be obtained for all other uses, in any current or future media, including reprinting/republishing this material for advertising or promotional purposes, creating new collective works, for resale or redistribution to servers or lists, or reuse of any copyrighted component of this work in other works. \hfill }
%		\hspace{\columnsep}\makebox[\columnwidth]{\hfill }}
	
	\title{Adversarial Attacks on Remote User Authentication Using Behavioural Mouse Dynamics
	}
	
	\author{\IEEEauthorblockN{Yi Xiang Marcus Tan,\IEEEauthorrefmark{1}\IEEEauthorrefmark{3} Alfonso Iacovazzi,\IEEEauthorrefmark{1} Ivan Homoliak,\IEEEauthorrefmark{1} Yuval Elovici,\IEEEauthorrefmark{1}\IEEEauthorrefmark{2} and Alexander Binder\IEEEauthorrefmark{1}\IEEEauthorrefmark{3}}\\
		\IEEEauthorblockA{\IEEEauthorrefmark{1}ST Engineering Electronics-SUTD Cyber Security Laboratory\\
			\IEEEauthorrefmark{3}Information Systems Technology and Design (ISTD) Pillar, Singapore University of Technology and Design, Singapore\\\
			%		Singapore University of Technology and Design, Singapore\\
			\IEEEauthorrefmark{2}Department of Software and Information Systems Engineering, Ben-Gurion University of the Negev, Beer-Sheva, Israel\\
			\IEEEauthorrefmark{2}Deutsche Telekom Innovation Laboratories at Ben-Gurion University of the Negev, Beer-Sheva, Israel\\
			Email: marcus\_tan@mymail.sutd.edu.sg\\
			Email: \{alfonso\_iacovazzi, ivan\_homoliak, yuval\_elovici, alexander\_binder\}@sutd.edu.sg
	}}

	\maketitle

	\begin{abstract}

		Mouse dynamics is a potential means of authenticating users. Typically, the authentication process is based on classical machine learning techniques, but recently, deep learning techniques have been introduced for this purpose. Although prior research has demonstrated how machine learning and deep learning algorithms can be bypassed by carefully crafted adversarial samples, there has been very little research performed on the topic of behavioural biometrics in the adversarial domain. In an attempt to address this gap, we built a set of attacks, which are applications of several generative approaches, to construct adversarial mouse trajectories that bypass authentication models. These generated mouse sequences will serve as the \textit{adversarial samples} in the context of our experiments. We also present an analysis of the attack approaches we explored, explaining their limitations. In contrast to previous work, we consider the attacks in a more realistic and challenging setting in which an attacker has access to recorded user data but does not have access to the authentication model or its outputs. We explore three different attack strategies: 1) statistics-based,
		2) imitation-based, and 3) surrogate-based; we show that they
		are able to evade the functionality of the authentication models,
		thereby impacting their robustness adversely. 
		%We show that imitation-based attacks often perform better than surrogate-based attacks, unless, however, the attacker can guess the surrogate-based attack model architecture, which we propose a detection mechanism for.
		We show that imitation-based attacks often perform better than surrogate-based attacks, unless, however, the attacker can guess the architecture of the authentication model. In such cases, we propose a potential detection mechanism against surrogate-based attacks. 
		%We show that imitation-based attacks often perform better than surrogate-based attacks, which involves the act of training a substitute model. However, if the attacker guesses the architecture of the authentication model, surrogate-based attacks would perform better. In such cases, we propose a potential detection mechanism against surrogate-based attacks.

		%, even with simple adversarial attack techniques. 

		%\todo{Add more interesting conclusions heere}
		%NEED TO ADD SOME RESULTS HERE
	\end{abstract}
	
	%\begin{IEEEkeywords}
	%Adversarial Attacks, Mouse authenticators, Machine Learning, Sequence generation
	%\end{IEEEkeywords}

	\section{Introduction}
	\label{intro}
	\IEEEoverridecommandlockouts
	\IEEEpubid{\begin{minipage}{\textwidth} \mycopyrightnotice
	\end{minipage}}
	\IEEEpubidadjcol
	\IEEEpubidadjcol
	%, whether in the classical or deep learning approach
	Static authentication (e.g.,~passwords and PINs) has been the predominant means of performing user authentication in many computer systems. With the demonstrated effectiveness of machine learning methods, researchers have turned to these methods to perform biometrics-based authentication, both physiological \cite{bhattacharyya2009biometric} (e.g.,~iris, facial recognition and fingerprints) and behavioural (e.g,~keystrokes \cite{monrose2000keystroke} and mouse dynamics). Although physiological-based authentication has proven to be highly effective, it can be costly to implement due to dedicated hardware requirements. 
	Furthermore, this form of authentication is easily bypassed if one can have access to a copy of the required features, since the features are usually clearly defined.
	Behaviour-based authentication, on the other hand, has the potential to be much more cost efficient, since it does not require extra hardware to accomplish the same task. Moreover, its non-intrusive nature allows users to be authenticated continuously, which provides an additional layer of security by only allowing the legitimate user to have access and continuous usage of a protected resource.
	
\begin{figure*}
		\centering
		\begin{subfigure}[]{0.33\textwidth}
			\centering
			\includegraphics[trim={0.8cm 0 0.8cm 1.35cm},clip, width=0.7\textwidth]{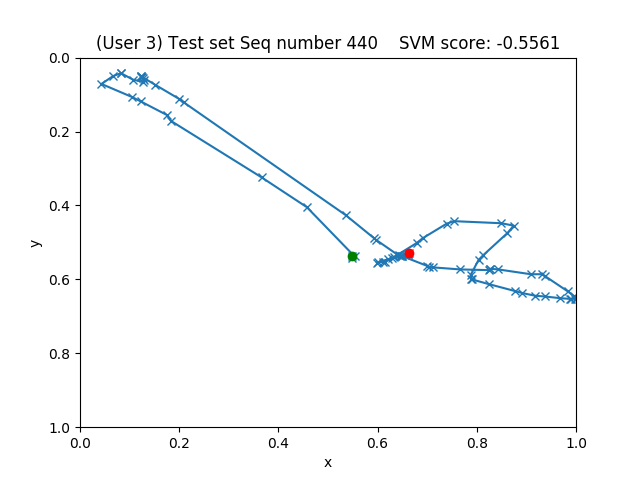}
			\caption{Sample mouse sequence of User 3}
			\label{fig:user3}
		\end{subfigure}%	
		\begin{subfigure}[]{0.33\textwidth}
			\centering
			\includegraphics[trim={0.8cm 0 0.8cm 1.35cm},clip, width=0.7\textwidth]{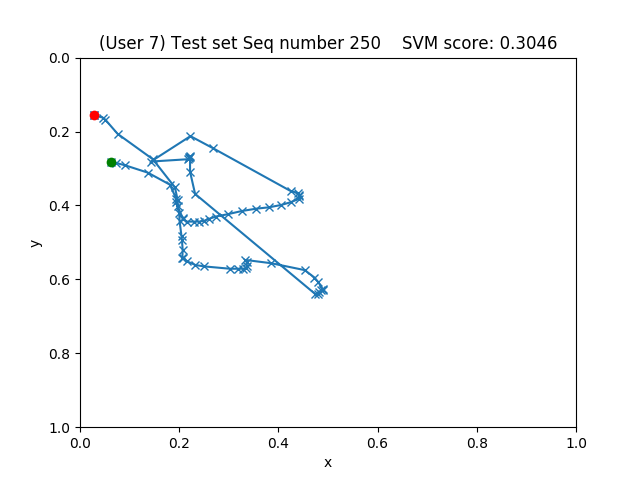}
			\caption{Sample mouse sequence of User 7}
			\label{fig:user7}
		\end{subfigure}%
		\begin{subfigure}[]{0.33\textwidth}
			\centering
			\includegraphics[trim={0.8cm 0 0.8cm 1.35cm},clip, width=0.7\textwidth]{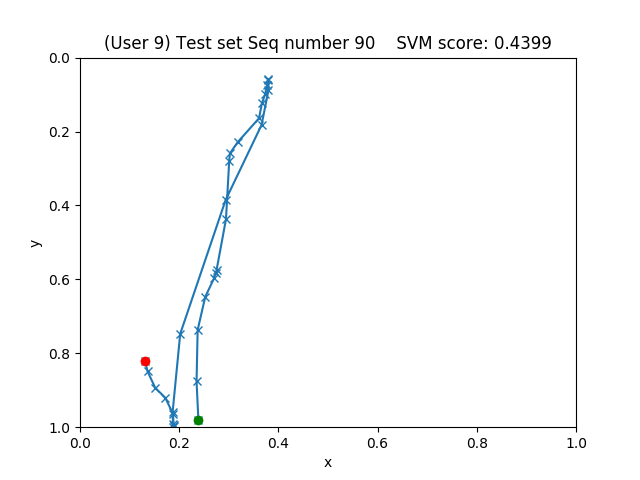}
			\caption{Sample mouse sequence of User 9}
			\label{fig:user9}
		\end{subfigure}%
		\caption{Sample mouse sequences belonging to three different users from the Balabit dataset \cite{balabit} illustrating the relative difficulty of the authentication problem. This is because human visual inspection does not reveal whether a sample contains information relevant to the prediction of the legitimacy of a particular user, in contrast to many problems in the image domain for which adversarial attacks have been applied (e.g., digits and object classification).		}\label{fig:seqs}
	\end{figure*}

	In the area of behaviour-based authentication by means of mouse dynamics analytics, a number of efforts have been undertaken \cite{Ahmed2007,Chong2018,Feher2012,Gamboa2004,Kasprowski2018,Mondal2017,Shen2012,Shen2013,Sayed2013,Mo2018,5291887,5337543,Zheng2011,  monrose2000keystroke} to improve the performance of mouse dynamics authentication models. They range from using statistics-based approaches \cite{Gamboa2004, Mondal2017}, classical machine learning models \cite{Feher2012, Kasprowski2018}, and to using deep learning \cite{Chong2018} to perform authentication tasks. 
	%However, to the best of our knowledge, very little work has been done in assessing the robustness of those models to adversarial attacks. 
	Also, most of the adversarial attacks proposed in prior studies focused on the image domain \cite{Papernot2016, Papernot2016a, Goodfellow2014, carlini2017towards,brendel2017decision,evtimov2017robust, szegedy2013intriguing}, which is considered less challenging since there are distinct and observable features within images that humans can identify. 
	%Furthermore, images potentially have a greater flexibility for performing perturbations in the input space as compared to sequential data due to its higher input dimensionality.
	%Furthermore, a human can visually verify the plausibility of any results. 
	In contrast, mouse sequences are hard to distinguish through the mere act of visual inspection by humans, as illustrated by Figure \ref{fig:seqs}.
	%This point is illustrated in Figures \ref{fig:user3},  \ref{fig:user7} and \ref{fig:user9}, where the three curves show sample mouse trajectories from three different users.
	Therefore, to the best of our knowledge, very little work has been done in the area of performing adversarial attacks in the domain of mouse dynamics, which deals mainly with temporal data, and also assessing the robustness of those models to such attacks and we attempt to address these gaps in this work.
	\IEEEpubidadjcol
	\IEEEpubidadjcol
	We consider a more realistic attack scenario than in previous works\cite{Papernot2016, Papernot2016a, Goodfellow2014, carlini2017towards,brendel2017decision,evtimov2017robust, szegedy2013intriguing}; in our attack scenario an attacker has access to the victim's host machine (i.e. client), but does not have access to the authenticators or their outputs which are deployed on a remote secured central authentication server. The attacker is only capable of recording the mouse movements of the targeted user. 
	%We make the assumption of a remote server because such verification is most likely to be done in highly secured environments (systems involving critical infrastructure), in order to provide the required level of security. As such, we consider a setting of this nature in this work. 
	This makes the attack scenario more challenging compared to scenarios in which the attacker has some kind of access to the authenticators.
	
	The contributions of our research include the following:
	% focus on the questions to answer here instead of writing what is done
	\begin{enumerate}
		\label{contrib}
		\item Considering the remote behavioural mouse dynamics
		authentication scheme, we propose three adversarial
		approaches to evade the authentication models:
		statistics-based, imitation-based, and surrogate-based. We examine this in a novel and realistic setting as stated above.
		% 	\item \label{contrib:point1} We consider a more realistic attack scenario compared to previous works, where an attacker has obtained access to the computer of a targeted user, but not having access to the authentication models or their outputs which are deployed on a secured central authentication server. The attacker is only capable to record the behavioural traits of the targeted user. Figure \ref{threatmodelgraphic} illustrates this point. This makes the attack scenario more challenging compared to approaches that have a some kind of access to the authenticator models.
		
		\item We address the question of which approach is the most effective for performing adversarial attacks in this kind of setting. We evaluate the robustness of machine learning-based authentication models to such adversarial attacks. As representative examples, we choose one authenticator based on engineered features and another based on neural networks.
		\item We analyse the relationships between the difficulty of the authentication task, the type of adversarial attack, and the adversarial success rate. %We observe a negative correlation between the hardness of the authentication task and the success rate of adversarial attacks.
		\item Given that the attacker is able to guess the architecture
		of the models, we perform the surrogate-based attack
		and analyse its impact on the robustness of the models.
		Finally, we propose a potential detection mechanism.
	\end{enumerate}
	%Also, we would like to reiterate the main research question we would like to answer in this work, which is to gain an understanding of the level of confidence we can place in such behavioural-based authenticator models in the domain of mouse dynamics.
	
	%In our work, we focus on attempting to fool authentication models, based on some assumptions made about our threat model, which will be discussed in Section \ref{threatmodel}. In the context of our experiments, we define \textit{adversarial samples}, which are mouse sequences not generated by a human, being labelled as legitimate as a successful attempt in fooling a targeted classifier model. Also, we would like to reiterate the main research question we would like to answer in this work, which is to gain an understanding of the level of confidence we can place in such behavioural-based authenticator models in the domain of mouse dynamics.
	%\todo{add in research qustions to answer?}

	This paper is organized as follows. 
	Section \ref{related} discusses related prior work, while Section \ref{threatmodel} lays out our assumptions regarding the attacker and provides an example use case. 
	In Section \ref{method}, we introduce our adversarial attack approaches. 
	In Section \ref{exp}, we present our experimental results; 
	this is followed by a discussion of the results in Section  \ref{discuss} and our conclusions in Section \ref{conclude}.
	
	\section{Related Work}
	\label{related}
	%\subsection{About Mouse Dynamics}

	\subsection{Mouse Dynamics Authentication Models}
	%One of the early works that was focused in this domain was \cite{Gamboa2004}. The authors extracted handcrafted features from mouse movement sequences, generated from a cubic spline interpolation preprocessing step, and then used statistical classifiers to perform the user verification task. 
	%One of the early works focusing in this domain was \cite{Gamboa2004}
	The authors in \cite{Gamboa2004} defined a total of $63$ handcrafted features that represents the mouse movement sequences' spatial and temporal information. However, in this research, the authors did not use the full set of features but instead, performed a feature selection step that is user-specific in order to obtain a subset of the most ``discriminative'' features. The resultant subset was then used for training statistics-based classifiers tailored to each user.
	%In \cite{ahmed2007new}, the authors considered 3 different types of mouse actions which were Point and Click (PC), Drag Drop (DD) and Mouse Movement (MM). Manual feature extraction was performed on these actions, extracting a total of $39$ features. These input features were fed into a Multi-Layered Perceptron that has a single hidden layer, which was trained to give a binary-labelled prediction indicating the authenticity of the user in question.
	The authors in \cite{Feher2012} used multi-level feature aggregation based on the features introduced in \cite{Gamboa2004}, which concatenates various mouse actions to form higher level features. Furthermore, the authors proposed additional features that were derived from individual mouse actions in order to reduce authentication time, due to the aggregation of several mouse sequences, while also improving the authentication accuracy. The final feature set was then used to train a random forest classifier model that learns to discriminate between a legitimate and an illegitimate users. 
	% The work done in \cite{Feher2012} uses multi-level feature aggregation, based on the features introduced in \cite{Gamboa2004} and their own proposed features, that concatenates various mouse actions to form higher level features. The proposed features were derived from individual mouse actions to reduce verification time, due to the aggregation of several mouse sequences, while also improving the verification accuracy. The final feature set was then used to train a Random Forest classifier model that learns to discriminate between a legitimate and an illegitimate user. 
	
	More recently, the authors in \cite{Chong2018} utilized a deep learning approach, more specifically a two-dimensional convolutional neural network (2DCNN), to learn the characteristics of users' mouse sequences for authentication. They gathered mouse dynamics sequences and converted them to images of the mouse trajectories. The authors showed that even by removing the temporal aspect of mouse movements, they still could achieve state-of-the-art authentication results on different mouse datasets. Furthermore, in \cite{Chong2018}, explanations were derived regarding which parts of mouse sequences contain evidence for a particular user by layer-wise relevance propagation \cite{binder2016layer, bach2015pixel}.
	
	%Furthermore, the authors went on to perform Layer-wise Relevance Propagation (LRP) \cite{binder2016layer} to explain how the model arrived at its decision.
	%\textit{(whether the input image came from a legitimate user or not)}.
	
	%They then used the 2DCNN to train on these trajectory images, to learn 
	
	%to perform mouse dynamics authentication task, achieving state-of-the-art results.
	
	\subsection{Adversarial Attacks}
	%In recent years, adversarial machine learning has garnered much attention by researchers
	Adversarial attacks against machine learning classifier models can be categorized into white-box or black-box attacks. For white-box attacks, it is assumed that the attacker has full access to a fully-differentiable target classifier \textit{(weights, architecture, and feature spaces)}. This allows the attacker to perform gradient-based attacks, for instance the fast gradient sign method (FGSM), which was proposed by the authors in \cite{Goodfellow2014}. This method involves calculating the gradients of the loss with respect to the model inputs, and subsequently, using it to perturb the input sample, in order to bring the resultant perturbed sample closer to the point of being misclassified. In addition to presenting the FGSM, in \cite{Papernot2016a} the authors proposed the jacobian-based saliency map attack (JSMA). 
	% It can be thought of as a more refined manner of performing perturbations on the input samples, as 
	The JSMA approach first calculates the impact of changing a particular feature (originally from class $j$, to a target class $i$) on the predicted label. Features of higher significance will be perturbed up to a defined threshold. These two proposed methods proved to be highly effective in attacking the robustness of the victim's machine learning model. However, having white-box access to a victim's machine learning model is highly unrealistic, as highly sensitive information, such as the model's weights and architecture would be a well-guarded secret in practice. % and not being stored on the machine of the targeted user. %Also, it can be expected that the authentication model runs on a separate server and obtaining access to the server implies that there will be other methods available to prevent the authentication model to be effective such as directly overwriting the model outputs. 

	%it can be expected that the authentication model runs on a separate server and obtaining access to the server implies that there will be other methods available to prevent the authentication model to be effective such as directly overwriting the model outputs
	% \textbf{(TODO refine this explanation)}
	
	In \cite{Papernot2016}, the authors addressed this limitation by introducing the concept of performing black-box attacks against machine learning classifiers, both deep neural networks (DNNs) and classical-based methods as well. 
	% The authors relax the assumption that attackers have access to the internals of the target classifier. 
	The authors require only the ability to query the classifier with a custom sample and to access its outputs. They performed their attack by training a surrogate classifier on a synthetic dataset which is labelled by the target classifier by sending sample queries and collecting its response. It should be noted that the requirement here is to be able to feed in inputs and to collect outputs from the authenticator. The authors then showed that the trained surrogate model is able to approximate the decision boundary of the target classifier, providing the opportunity for white-box attacks.
	The authors concluded by showing that attacking the robustness of the victim's model is feasible, even with defences in place during the phase of training the victim models. It is important to note that the context of the work done by authors in \cite{Papernot2016, Goodfellow2014, Papernot2016a} was in the image domain.
	
	%Talk about Yuval paper here.
	%Extending the attack methodology proposed in \cite{Papernot2016} to another separate domain,
	In \cite{Rosenberg2018}, the authors adopted the methodology used in \cite{Papernot2016} and applied it to perform adversarial attacks on malware classifiers. Similar to \cite{Papernot2016}, the authors trained a surrogate malware classifier and generated malware through the method proposed by \cite{Goodfellow2014}. The malware resulting from the perturbation was still able to fulfil its original malicious functionality due to the introduction of ``No Operations'' (NOPs) as a form of perturbation. The authors achieved nearly perfect rates of bypassing the targeted malware classifiers with this approach.
	
	However, none of the attacks described above were applied against machine learning-based authentication models. Furthermore, in contrast to the aforementioned adversarial attacks, in the domain of machine learning-based authentication, it is not realistic to assume that the adversary has access to the authentication model; in this case, the model might be situated in a remote server, away from the victim computer given a highly secured environment.
	%due to the fact that from a system security perspective, it is not realistic to assume that the verification model is stored on the host machine of the victim. 
	Hence, the attacker would not have any knowledge of the outcome of the authentication computed by this remote server and thus be unable to perform black-box attacks proposed by the authors in \cite{Papernot2016}. For the same reason, white-box attacks that rely on access to the architecture and weights of the model are not possible. 
	\section{Threat Model}
	\label{threatmodel}
	%Talk about the assumptions made about the attacker. What information does the attacker have?
	%In this section we discuss the realistic assumptions we make about the attacker, comprising of the attacker's goals, capability and also knowledge of the system.
	
	While the focus of this work is on adversarial attacks, we first provide a justification for the considered setup by an example use case before describing our threat model proper. We consider a remote user authentication scenario, based on mouse behavioural dynamics in the setting of a client/server architecture (see Figure 2). In this setting, a user on the client machine is allowed to access a remote service on a server if the user’s currently collected behavioural characteristics are consistent with the model stored at the remote server. On the other hand, if the user’s machine is compromised by an external attacker (e.g., by exploiting a vulnerability or installing a malware) or by an internal attacker (i.e., insider threat such as masquerader \cite{homoliak2018}), then after a sufficient period of time, a record about this activity would be generated at the server. Additionally, the session of the user might get interrupted by additional verification measures.
	
	%While the focus of this paper is on attacker models, lets give at first a justification for the considered setup by an exemplary use case. We consider remote user authentication based on mouse behavioural dynamics in the setting of a client/server architecture (see Figure 2). In this setting, a user on a client PC is allowed to consume a remote service on a server if the user’s currently collected behavioural characteristics are consistent with the model stored at the remote server (i.e., target classifier). On the other hand, if the user’s machine is compromised by an external attacker (e.g., by exploiting a vulnerability or installing a malware) or by an internal attacker (i.e., insider threat such as masquerader [25]), then after a sufficient period of time, a record about this activity generated at the server. Additionally, the session of the user might get interrupted by additional verification measures.

\begin{figure}[tbp]
		\centering{\includegraphics[trim={0.0cm 0 1.6cm 0.0cm},clip,width=0.6\linewidth]{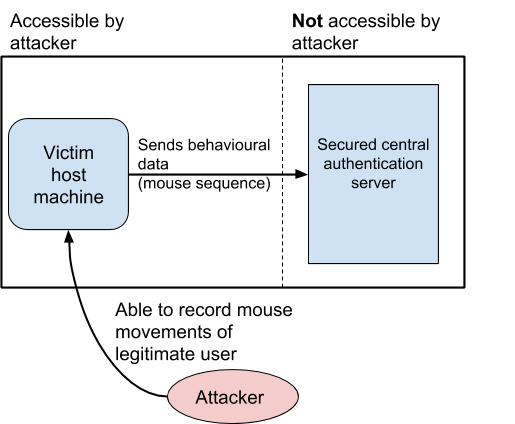}}
		\caption{Illustration of the proposed threat model.}
		\label{threatmodelgraphic}
	\end{figure}

	% As mentioned in Section \ref{intro}, t
	The attacker's goal is to have the generated adversarial samples evade a machine learning-based authentication model (i.e. target classifier). We assume that the attacker does not target the secured central authentication server (see Figure \ref{threatmodelgraphic}) but instead compromises the host machine of the target user. Thus, the attacker has the ability to record the mouse movement sequences of the legitimate user on the infected host machine. 
	%In \cite{Papernot2016}, the authors assumed that the attacker has access to the target classifier by querying it for its predictions, although they assumed that the target classifier was a black-box. 
	We extend the threat model assumed by the authors in \cite{Papernot2016}, 
	by not allowing the attacker to query the target classifier, which
	makes our threat model more realistic and at the same time
	more challenging.
	%In our study, 
	%we take this further by assuming that we do not have the ability to query the target classifier. This makes our threat model more realistic but at the same time more challenging. 
	It is important to note here that the recorded data is not a subset of the training data used to train the target classifier and also, no feedback regarding the result of the authentication will be accessible to the attacker. 
	%Figure \ref{threatmodelgraphic} summarizes our assumptions regarding the attacker. 
	If the attacker's goal is achieved to a reasonable extent, the robustness of target classifiers can indeed be adversely affected and the reliability of these models must be verified, regardless of their authentication performance. 
	
	\section{Proposed Adversarial Strategies}
	\label{method}
	%In this work, our objective is simply to fool a behavioural-based target classifier.
	%In this work, our objective is to evaluate the robustness and reliability of behavioural-based target classifiers. We experimented with $2$ different approaches to achieve our goal.
	
	%In this work, our objective is to evaluate the robustness and reliability of behaviour-based target classifiers. For this purpose, we built a set of attacks and assessed their effectiveness. 
	We investigated three possible attack approaches used to bypass the target classifiers. The first approach trains a generator model to impersonate a user's mouse movement sequence through the teacher-forcing approach \cite{Williams1989}, which we refer to as an imitation-based attack. This trained model then generates mouse trajectories, which will be tested against the target classifier. 
	The second method is based on the idea of training a surrogate classifier, as proposed in \cite{Papernot2016}, however, while assuming that the attacker has no access to the target classifier. We refer to this method as a surrogate-based attack. This trained surrogate model will be used to perform perturbation of any mouse trajectory, through the use of a white-box attack method. The perturbed sequences will then be evaluated against the target classifier. Hence, our first method involves the generation of mouse curves, while the second involves the perturbation of any existing mouse curves. 
	Finally, these two methods are compared against a third attack approach, which we refer as a statistics-based attack. 
	%The details of the statistics-based attack are discussed in Section \ref{statsbased}.
	
	\subsection{Statistics-based Attack}
	As a baseline, we adopted a statistical-based approach to generate adversarial trajectories in an attempt to bypass the target classifier as it was the simplest and fastest to implement. It does not require any overhead in training any neural network models for performing such adversarial attacks. As mentioned earlier in Section \ref{threatmodel}, we assumed that the attacker has access to recorded mouse dynamics of the target user; 
	hence, the attacker can calculate several useful statistics of the user. 
	%hence, the attacker can extract useful information that represents the user
	For sequence generation, we first compute a histogram of position difference vectors, a histogram of starting points, and the median time interval between each mouse event. We then select bins in the histograms through random sampling, weighted on the rate of occurrence. As a bin contains a range of possible values, we sampled uniformly within the bounds of the selected bin, in order to obtain a mouse coordinate. From the histogram of position difference vectors, we were able to sample a sequence of mouse coordinate perturbations. With the sampled start positions (from the histogram of starting points), each subsequent point can be generated by adding the perturbation from the last computed position vector to form a sequence of absolute position vectors. The median time interval was then used to construct timestamps for each generated mouse event.
	
	\subsection{Imitation-based Attack}
	\label{method:teacher}
	%We used this variant instead of the Long-Short Term Memory (LSTM) variant due to its lower computational cost. Similar to the LSTM variant,
	
	We employed the gated recurrent unit variant of recurrent neural networks (GRU-RNN) architecture as the basis of our generator network, $g$. The GRU has the ability to learn what to forget through the reset gate, $r^{(t)}$, as well as the relations between data points across timesteps in a sequence. The governing equations for a GRU cell are shown below:
	%\begin{align}
	%\label{grucell:1}
	%r_t &= \sigma(W_{ir}x_t + b_{ir} + W_{hr}h_{t-1} + b_{hr})\\
	%\label{grucell:2}
	%z_t &= \sigma(W_{iz}x_t + b_{iz} + W_{hz}h_{t-1} + b_{hz})\\
	%\label{grucell:3}
	%n_t &= \tanh(W_{in}x_t + b_{in} + r_t(W_{hn}h_{t-1} + b_{hn}))\\
	%\label{grucell:4}
	%h_t &= (1-z_t)n_t + z_th_{t-1}
	%\end{align}
	%\begin{align}
	%\label{grucell:1}
	%r^{(t)} &= \sigma(W_{ir}x^{(t)} + b_{ir} + W_{hr}h^{(t-1)} + b_{hr})\\
	%\label{grucell:2}
	%z^{(t)} &= \sigma(W_{iz}x^{(t)} + b_{iz} + W_{hz}h^{(t-1)} + b_{hz})\\
	%\label{grucell:3}
	%n^{(t)} &= \tanh(W_{in}x^{(t)} + b_{in} + r^{(t)}(W_{hn}h^{(t-1)} + b_{hn}))\\
	%\label{grucell:4}
	%h^{(t)} &= (1-z^{(t)})n^{(t)} + z_th^{(t-1)}
	%\end{align}
	\begin{align}
	\label{grucell:1}
	r^{(t)} &= \sigma(W_{ir}X^{(t)} + b_{ir} + W_{hr}h^{(t-1)} + b_{hr})\\
	\label{grucell:2}
	z^{(t)} &= \sigma(W_{iz}X^{(t)} + b_{iz} + W_{hz}h^{(t-1)} + b_{hz})\\
	\label{grucell:3}
	n^{(t)} &= \tanh(W_{in}X^{(t)} + b_{in} + r^{(t)}(W_{hn}h^{(t-1)} + b_{hn}))\\
	\label{grucell:4}
	h^{(t)} &= (1-z^{(t)})n^{(t)} + z^{(t)}h^{(t-1)}
	\end{align}
	
	\noindent where $W$ and $b$ are the weights and biases respectively, and $h^{(t)}$ is the output of the GRU cell at timestep $t$. $\sigma$ is the sigmoid activation function. 
	% that limits its outputs to between 0 and 1.
	
	% For the remaining of the paper we refer to this model as the generator, $g$.
	
	The GRU-RNN model was trained to predict the coordinates of the next timestamp, $\hat{X}^{(t+1)}$, given the ground truth from the current timestamp, $X^{(t)}$, and the hidden states of the previous timestamp, $h^{(t-1)}$. It can be expressed as:
	\begin{equation}
	\label{rnneqn}
	\hat{X}^{(t+1)},h^{(t)}=g(X^{(t)}, h^{(t-1)};\theta_g)
	\end{equation}
	\noindent where $\theta_g$ denotes the model parameters of the generator network $g$.
	%\noindent where $\hat{x}^{(t+1)}$ is the predicted data point in the next timestamp, $x^{(t)}$ is the current timestamp's data point, $h^{(t-1)}$ is the hidden state outputs of the previous timestep and $\theta_g$ denotes the model parameters of the network $g$.
	%During training, the model seeks to maximize the likelihood of predicting the correct token $\hat{x}^{(t+1)}$
	The prediction of the generator, $\hat{X}_i^{(t+1)}$, and the ground truth, $X_i^{(t+1)}$, are used to calculate the mean square error (MSE) loss, which will be used for backpropagation to update the model parameters, $\theta_g$. 
	
	%The loss function is defined as:
	%\begin{equation}
	%\label{mseloss}
	%L(X^{(t)}, \hat{X}^{(t)}) = \frac{1}{N}\sum_{i=0}^{N}(X_i^{(t)}-\hat{X}_i^{(t)})^2
	%\end{equation}
	%\noindent where $i$ is the sample index.
	%Figure \ref{imitationlearn} shows an illustration of the imitation-based generator training. Please note that the GRU layer can be extended to multiple GRU layers.
	
	%\begin{figure}[htbp]
	%	\centering{\includegraphics[scale=0.405]{imitationbased.png}}
	%	\caption{Simple example of Imitation-based generator training. Loss values are defined by L.}
	%	\label{imitationlearn}
	%\end{figure}
	Our GRU-RNN model consists of two stacked GRU layers, with a hidden dimension of 128, and a fully-connected layer at the end of the GRU sequence to convert a dimension of 128 to two (to translate the latent space back to an x-y space).
	
	There are a number of design choices for the representation of $X$, and in our experiments, we used three different variants. They are absolute position vectors (ABS), position difference vectors (DVs) between the mouse coordinates of the current and the previous timestamp, and, velocities (VELs) which are DVs divided by the corresponding time difference. In addition, we have experimented training our generator with and without regularization.
	Generating mouse movement sequences was done in two ways, by giving the generator an initial single start position for the generator model to start generating a full length sequence and by providing an initial sequence of coordinates, taken from the recorded mouse trajectories, to start the generation process. More details regarding the generation of these sequences are provided in Section \ref{imitateexp}.
	%\todo{Add in the other regularizers etc. Have cluster reg and vel/acc reg}
	
	%Also, we have also tried training the Generator with and without regularization. The variations we made are as follows:
	%\begin{enumerate}
	%	\item ABS with velocity regularizer
	%	\item DIFFVEC with velocity regularizer
	%	\item VEL with acceleration regularizer
	%\end{enumerate}
	
	\subsection{Surrogate-based Attack}
	\label{method:surrogate}
	\begin{figure}[tbp]
		\centering{\includegraphics[width=0.4\textwidth]{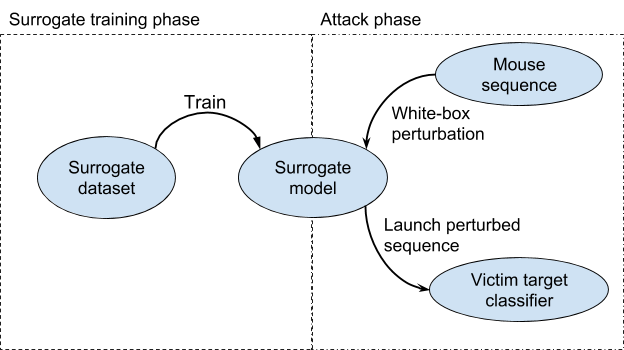}}
		\caption{Workflow of a surrogate-based attack}
		\label{surrogatefool}
	\end{figure}

	We trained a surrogate classifier, inspired by \cite{Papernot2016}, that learns to have the same functionality as the target classifier from a surrogate dataset, even though their network architectures might be different. This requires positive and negative labelled samples for training. Therefore, the surrogate dataset is comprised of the target user's mouse movement sequences recorded by the attacker and a set of other mouse sequences that do not belong to the user. %For simplicity, the class labels that we used for the surrogate dataset are binary where ``1'' refers to mouse sequences coming from the victim user while ``0'' refers to any other mouse movement sequences not belonging to the victim. 
	After training the surrogate classifier, we performed white-box attacks on the surrogate model, by starting with an arbitrary mouse movement sequence and perturbing it for a fixed number of iterations. %, until it is classified as legitimate under the surrogate. 
	In order to accomplish this, we adopted the FGSM \cite{Goodfellow2014}.
	%Let the surrogate model be $S$, its model parameters be $\theta_S$, $z$ our input mouse trajectory to the model $S$, $y$ as the intended class $z$ should become after perturbation, and $J_S(S(z), y, \theta_S)$ be the loss function used when training the surrogate model. 
	The amount of perturbation, $\delta z$, to be applied on the input sequence, $z$, can be defined as:
	\begin{equation}
	\label{fgsm}
	\delta z=\epsilon \cdot sign(\nabla_z (\log p(z) ))
	\end{equation}
	\noindent where $\epsilon$ is the perturbation factor that controls the extent of the perturbation, and $\log p(z)$ is the log probability of the legitimate user class for the input sequence, $z$. Note that the surrogate model can have any architecture and not necessarily have the same architecture as the target classifier. Also, we used the cross entropy loss function. Figure \ref{surrogatefool} illustrates the process of the described surrogate-based attack.
	
	We experimented with two different surrogate model architectures. One was a three-layer GRU-RNN with two fully-connected (FC) layers on top and a rectified linear unit (ReLU) activation after the first FC layer. The model has a hidden dimension of 100, and a fully-connected layer at the end to convert the GRU-RNN outputs to two-dimensional vectors, which represents logits for each decision (zero or one).
	The second surrogate only uses two stacked FC layers, with an exponential linear units (ELU) activation after the first layer. The first layer takes in a flattened sequence as inputs, where every \textit{odd} numbered node takes in the x-axis, while every \textit{even} numbered node takes in the y-axis of the mouse sequence.\footnote{For example, if a mouse sequence has a sequence length of 10 (each timestep being represented as a vector consisting of x and y coordinates), then the input dimension of the first layer would be 20.} The output of the second model has the same format as the first. 
	%For both architectures, the outputs will be passed to a \textit{Cross-Entropy loss function} for loss calculation and eventual backpropagation. The FC surrogate model has hidden dimensions that are halved from the input dimensions at each FC layer except the last, where the output dimensions would be 2.
	%We considered two different architectures for our surrogate models. The GRU-RNN model has a hidden dimension of 100, and a fully-connected layer at the end to convert the GRU-RNN outputs to 2-dimensional vectors. 
	%The FC surrogate model has hidden dimensions that are halved from the input dimensions at each FC layer except the last, where the output dimensions would be 2.
	In contrast to the work of \cite{Papernot2016}, we do not use the target classifier to label our samples, due to our assumptions mentioned in Section \ref{threatmodel}. 
	% Since we do not have access to the target classifier model.%, we do not consider the transferability property between the surrogate model and target classifier models that \cite{Papernot2016} mentioned. 

	%%%%%%%%%%%%%%%%%%%%%%%%%%%%%%%%%%%%%

	\section{Experiments}
	\label{exp}
	%In order to evaluate the robustness of authenticator models to adversarial attacks, and also generalize it to different kinds of datasets, we use two different publicly available mouse datasets.
	In order to comprehensively evaluate the effectiveness of the adversarial attacks and also the robustness of the target classifiers to adversarial attacks, we used two different publicly available mouse datasets.
	In our experiments, we used the PyTorch library \cite{paszke2017automatic} for developing our models unless stated otherwise.
	\subsection{Datasets}
	\label{datasets}
	%We used mouse dynamics datasets from Balabit \cite{balabit} and The Wolf of SUTD (TWOS) \cite{article}. 
	We used the Balabit \cite{balabit} and The Wolf of SUTD (TWOS) \cite{article} datasets. 
	In both datasets, we preprocessed the raw data logs into mouse movement sequences, removing mouse scroll and click events as they are not part of our scope. The data contain critical information like the event timestamp (in seconds), mouse position vector, and the user's identity. 
	
	%It is important to note that these anomalous sequences do not contain malicious intent but instead normal usage from another user.
	
	The Balabit dataset consists of mouse dynamics data from $10$ users which is split into training and testing sets, with the training set consisting of only the corresponding user's data. For the testing set, there is a mix of other users' data to simulate anomalous mouse dynamics sequences for comparison against the legitimate user. In our experiments, these anomalous sequences were not needed and were removed, since the true identity of those anomalous sequences was not known. 
	% This dataset contains the timestamps for each mouse event and also its horizontal $x$ and vertical $y$ coordinates. 
	Unfortunately, in this dataset the screen resolution of the users is not known, therefore we estimated their screen resolutions by calculating the maximum coordinates for each dimension and mapping it to a finite set of screen resolutions. 
	
	The TWOS dataset consists of mouse dynamics data from $24$ users, one of which was not considered because of its small sample size. The TWOS dataset provided screen resolutions of the users, in contrast to the Balabit dataset. However, we found that the TWOS dataset is more unclean as there exists multiple instances of repeated mouse events (with the same timestamp, x-y coordinates) which are considered as anomalies (Balabit has less of such occurrences).
	
	%\todo{Add in how the data splits are done on training the different models? The 4 splits}
	
	%While ensuring mouse data from a session file does not get splitted up, redistribute the combined data into two sets of approximately equal size.
	
	%\textit{(For the case of TWOS, where each user might not necessarily have many session files, a session file might potentially be further broken down to what we term it as ``mini-sessions'', where each mini-session is separated by a time difference of at least 2 hours. This is done to potentially prevent any two mouse movement sequences which were generated one after another to be present in both the ``train'' and ``test'' dataset. This is due to the fact that certain users in the TWOS dataset have merely 2-3 available session files.)}
	For both datasets, we performed the following data reshuffling techniques in order to obtain a training set for authentication model training and a disjoint training set for simulating the recording of mouse sequences from the legitimate user on the victim's machine, while ensuring that all users were represented in each set. The latter subset will be used either for surrogate or generator training.
	\begin{enumerate}
		\item Combine session files from both training and test folders.
		\item Redistribute the combined data into two sets of approximately equal size, all while ensuring that the mouse data within each session file does not get separated.\footnote{For the case of the TWOS dataset, where each user might not necessarily have many session files, a session file might be further broken down to what we refer to as ``mini-sessions'', where each mini-session is separated by a time difference of at least two hours. This is done to try to prevent any two mouse movement sequences which were generated one after another to be present in both the ``training'' and ``test'' datasets. This is required as certain users in the TWOS dataset may merely have 2-3 available session files.}
		%	\begin{enumerate}
		%		\item[a.] For the case of TWOS, where each user might not necessarily have many session files, a session file might potentially be further broken down to what we term it as ``mini-sessions'', where each mini-session is separated by a time difference of at least 2 hours. This is done to potentially prevent any two mouse movement sequences which were generated one after another to be present in both the ``train'' and ``test'' dataset. This is due to the fact that certain users in the TWOS dataset have merely 2-3 available session files.
		%	\end{enumerate}
		\item Further split each set into a ``training'' and ``test'' subset, of $80\%$ and $20\%$ of the dataset respectively. 
	\end{enumerate}
	%\noindent We ensured that all users were represented in each split. %After these preprocessing steps, we used the first set to train and validate the target classifiers while the second set was used to train and validate the generator and surrogate models.
	%\todo{mention number of samples per user used?}
%	we then filtered to use only mouse sequences that are too short
	
	Due to the anomalous mouse events encountered as described prior, we implemented a simple data cleaning step to remove such events. As such, some mouse sequences might be shortened. We then conducted a check to use only mouse sequences that are minimally of a certain length. Hence, the number of eligible mouse sequences will be reduced. In order to circumvent the issue of having little data, we performed a data augmentation step for data expansion. We did so by performing 10 different affine rotations within $\pm5$\textdegree (chosen at random) for each sequence.
	
%	We performed data augmentation for dataset expansion by performing 10 different affine rotations within $\pm5$\textdegree for each sequence.
	%ADDED
	We chose to perform affine rotations as this allows us to both generate more mouse sequences while also maintaining the user's mouse trajectory characteristics for each generated curve. Mouse sequence features like velocity, acceleration and curvature will be consistent across these generated curves with such a data expansion policy. We used 5\textdegree as it would provide us with sufficient variability.
	% clockwise and anti-clockwise 
	%about the center of the sequence. 
	We also used the screen resolutions of each user to normalize each mouse coordinate to the range of [0, 1].
	
	\subsection{Authentication Models}
	\label{authmodels}
	We considered two different types of authentication models: a neural network-based model using sequential data and one based on engineered features. More specifically, we used a one-dimensional convolutional neural network (1DCNN) and support vector machines (SVM) respectively. We considered the SVM as it gives us a simple baseline authentication model which can be trained stably with small sample sizes. Also, we chose to use these two types of authentication models as we would be able to observe if the same deductions can be drawn across highly different authentication models.
	\subsubsection{One-Dimensional Convolutional Neural Network}
	\begin{figure}[tbp]
		\centering
		\includegraphics[width=0.5\textwidth]{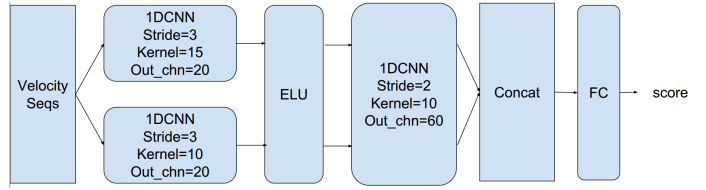}
		\caption{1DCNN architecture}
		\label{1dcnn}
	\end{figure}
	We choose a 1DCNN model as a neural network variant of our authentication model. This model was trained on sequences of velocities with a fixed sequence length which were preprocessed from sequences of position vectors with their corresponding timestamps. The model performs one dimensional convolutions along the time axis, with ELU as activation functions after each of the convolution layers. The first layer parses the input velocity sequence using two different time scales. The outputs of the first layer is passed to the second layer of convolution with shared weights. Lastly, the outputs of the second convolution layer are passed to a FC layer to produce a scalar value that represents the score of the input sample. The 1DCNN architecture is illustrated in Figure \ref{1dcnn}.

	%The 1DCNN model classifier was trained from scratch on our training data, after converting sequences of position vectors to sequences of velocities. The model takes in velocity sequences of $100$ timesteps
	
	%The model then perform 1 dimensional convolutions on each axis of the sequence of velocities to derive a final classification of whether the input velocity sequence came from the legitimate user or not.

	\subsubsection{Support Vector Machine}
	%We excluded two of the features from the set \cite{Feher2012} used such as Jitter and Critical Points
	
	%by filtering Jitter and Critical Points
	The SVM model was trained on $64$ features, adopted from the feature set that used in \cite{Feher2012}, however we used a slightly smaller feature space.\footnote{This was accomplished by filtering Jitter and Critical Points} We refer the reader to the ``Movement Features'' section for more information regarding the mouse movement features used. For data preprocessing, the input features were normalized by removing the mean and scaling it to unit variance.
	% before feeding it to the SVM model for training and evaluation. 
	The SVM model was developed using scikit-learn's LinearSVC API \cite{scikitlearn}. 
	%{Explain why it is removed, add appendix of table of features?} 
	
	%, which are widely used in biometrics
	The performance of the authenticators was measured by the area under curve\footnote{Receiver operating characteristic area under curve} (AUC) and equal error rate (EER) metrics. %The AUC measures the area under the Receiver Operating Characteristic (ROC) curve while the EER measures the rate of which the false positive rate and the false negative rate are equal. AUC ranges from $0$ to $1$, where higher means better. EER also has the same range as AUC but the lower the EER is the better.
	%ROC curves are especially meaningful for biometrics because they show the performance of the biometric system
	%on several operating points corresponding to different decisional
	%thresholds.
	%\textbf{(Include AUC and EER for SVM and 1DCNN here. State that the classifiers are reasonable enough to act as an authenticator)}
	\begin{table}[tb]
		\centering
		\caption{Baseline average AUC and EER for the respective authenticators on the different datasets. The metrics for the Balabit dataset are an average across $10$ users, while the metrics for the TWOS dataset are an average across 23 users. The state-of-the-art performance can be found in \cite{Chong2018} with a two-dimensional convolutional neural network (2DCNN) architecture.}
		\begin{tabular}{ccccc}
			& \multicolumn{2}{c}{Balabit} & \multicolumn{2}{c}{TWOS} \\  \hline
			Authentication models& AUC           & EER          & AUC         & EER         \\ \hline
			\multicolumn{1}{c}{SVM}   & 0.82590       & 0.23129      & 0.80485     & 0.24103     \\ \hline
			\multicolumn{1}{c}{1DCNN} & 0.87469       & 0.12531      & 0.77158     & 0.29242     \\ \hline
			\multicolumn{1}{c}{2DCNN \cite{Chong2018}} & 0.96       & 0.10      & 0.93     & 0.13     \\ \hline
		\end{tabular}
		\label{authbase}
		
	\end{table}
	It is important to note here that in our work, we are not interested in achieving the state-of-the-art performance for the user authentication task, but instead focus on evading authenticators. As such, our authentication performance is lower than the state-of-the-art reported in \cite{Chong2018}. We do not use the 2DCNN authenticator proposed by \cite{Chong2018} because it takes in input images instead of sequences. The purpose of Table \ref{authbase} is to show that the target classifiers that we used as our authentication models are reasonable and thus, that we are not attempting to attack a poor performing model. 
	%Also, the metrics reported under Balabit were averaged across $10$ users while for the case of TWOS, it was averaged across $23$ users.
	%that allows every mouse sequence to pass
	Based on Table \ref{authbase}, we can deduce that the authentication task is less complex for the Balabit dataset than it is for the TWOS dataset. 
	%has an easier verification task than TWOS. 
	The AUC for both authentication models using the Balabit dataset is higher than obtained using the TWOS dataset. Likewise, the EER for the Balabit dataset is lower than for the TWOS dataset. This behaviour is also supported by a prior work done by \cite{Chong2018}, with their state-of-the-art results as shown similarly in Table \ref{authbase}.
	
	\subsection{Adversarial Attack Results}
	In this work, we define the \textit{adversarial success rate (ASR)} as the proportion of \textit{generated/perturbed sequences} being \textit{classified as legitimate} by the targeted classifier, and thus having evaded the authenticator successfully.
	%\todo{talk about results of generation of seqs. Also compare with the stats based baseline}
	\subsubsection{Statistics-based Attack}
	\label{statsbased}

	Table \ref{statsbasedfooling} shows the ASR we obtained by using a statistics-based method of generating adversarial mouse trajectories, in an attempt to bypass the target classifiers described in Section \ref{authmodels}. With a simple statistics-based approach, target classifiers classified at least $26.4\%$ of the adversarial samples as legitimate, with a maximum of $61.54\%$ classified as legitimate. These results demonstrate that the target classifiers are not even robust to simple means of generating adversarial samples.
	
	\begin{table}[tbp]
		\centering
		\caption{Statistics-based attack results based on 1000 generated curves. Metrics reported represents the ASR obtained (in [0, 1])}
		\label{statsbasedfooling}
		\begin{tabular}{cccccc}
			&          & \multicolumn{2}{c}{Balabit} & \multicolumn{2}{c}{TWOS}    \\ 
			&    & SVM          & 1DCNN        & SVM          & 1DCNN        \\ \hline
			\multicolumn{2}{c}{Statistics-based baseline} & $0.6154 $      & $0.3183$       & $0.3895 $      &$ 0.264  $      \\ \hline
		\end{tabular}
	\end{table}
	%Using a simple statistical-based approach, the minimum adversarial success rate obtainable is 0.264. This shows that even with a non-neural network based method of generating adversarial samples, the target classifiers will still 

	\subsubsection{Imitation-based Attack}
	\label{imitateexp}

	Before describing our results, one should be aware of an inherent limitation of imitation-based attacks in the realistic scenario defined in Section \ref{threatmodel}. To illustrate this, consider a dataset and a method which trains a generator to accurately generate sequences according to the dataset perfectly. This generator would be able to replicate the instances from the dataset itself; as such, the ASR that the attacker can achieve would be equal to the accuracy of the authentication model, when evaluated on samples of the legitimate user. Thus, in the case in which the generator is able to reproduce the distribution of the provided dataset well, the ASR would be close to the accuracy of the authentication model on the legitimate sample but not significantly above
	%. There is no additional improvement expected on the generated sequences,
	as the authentication model cannot be used to improve the generator. 
	%It should be noted that when one cannot use the authenticator to guide the generator, this observation is expected to hold for any authentication method beyond mouse-based contexts, as the observation above is independent of the input modality. 
	It should be noted that when the authenticator cannot be used to guide the generator, this observation will likely hold for authentication methods beyond the mouse-based context, as the observation is independent of the input modality.
	% It also implies that the ASR of generative methods are in this weak sense tied to the accuracies of the authenticators. 
	% -- a successful generative method will be able to bypass . 

	% Increasing the variance of the generator may change the ASR, though without access to the authenticator, as in our setup, there is no improvement to be expected over this result on average over a set of generated sequences, as the authenticator cannot be used to improve the generator. This is a weak limit, which is expected to apply the more, 
	
	% the better the generator is able to learn the distribution of training sequences, and the lower the variance of generated samples is.

	\begin{table}[tbp]
		\centering
		\caption{Summary of setting variations used in the imitation-based attack experiments.}
		\label{imitationvar}
		\begin{tabular}{|c|c|l|c|c|c|l|}
			\hline
			\textbf{Generation Method}     & \multicolumn{3}{c|}{Start Point}        & \multicolumn{3}{c|}{Start Sequence}            \\ \hline
			\textbf{Model Strategy}        & \multicolumn{2}{c|}{ABS} & \multicolumn{2}{c|}{DV}      & \multicolumn{2}{c|}{VEL}        \\ \hline
			\textbf{Regularizing Strategy} & \multicolumn{2}{c|}{No}  & \multicolumn{2}{c|}{Cluster} & \multicolumn{2}{c|}{Derivative} \\ \hline
			\textbf{Sequence Length}       & \multicolumn{3}{c|}{50}                 & \multicolumn{3}{c|}{100}                        \\ \hline
		\end{tabular}
	\end{table}
	
	As mentioned in Section \ref{method:teacher}, the generator is trained using only the sequences of the victim and we also briefly described two methods of generating sequences, using a single start point or using a whole sequence\footnote{The generator's input sequence can be thought of as a queue. As more samples are being generated spanning across time, these generated samples are added to this queue, while pushing out the more outdated samples from the beginning of the sequence. } for initialization. In addition to varying the generation method, we have also experimented with training the generator with different regularization strategies and also with different sequence lengths. Table \ref{imitationvar} summarizes the different variations used in the imitation-based attack.
	In the Table, describing the regularization strategies, ``No'' refers to the lack of regularization during the training process. ``Derivative'' refers to regularizing each generated trajectory to the average of the user's velocity or acceleration, the former is used for ``ABS'' and ``DV'', while the latter is used for ``VEL''. ``Cluster'' refers to using a trained k-means clustering model based on a set of features of users described next. The regularization term is the Euclidean distance to the nearest cluster centroid in the feature space, where we used five clusters in our k-means. The features are statistics calculated from mouse sequences. We used the mean and standard deviation of the velocity (x and y direction), acceleration (x and y direction), and the angle of movement, resulting in a total of 10 features used in the clustering regularization method. 
	
	%\todo{Describe hyperparameters used to train model here}
	%The GRU-RNN model consist of 2 stacked GRU layers, with a hidden dimension of 128, and a fully-connected layer at the end of the GRU sequence to convert a dimension of 128 to 2 (to translate the latent space back to x-y space)
	
	The hyperparameters we used to train our generators are as follows. We used an Adam optimizer with a learning rate of 0.001. We adopted a learning rate decay strategy every 15 epochs by a factor of 0.5 and the models were trained for 60 epochs.

	Table \ref{imitationres2} summarizes the results obtained in the imitation-based attacks. Although we performed experiments enumerating all possible combinations of the settings described in Table \ref{imitationvar}, we aggregated the results over all model strategies and over all regularization strategies by computing the mean. This was done, because we found out that the effects of changes within these two settings did not produce any significant differences in the attacks' performance based on the Wilcoxon signed-ranked test \cite{wilcoxon1945individual}. More details about how this test was performed are discussed in Section \ref{wilcoxon}.
	
	\begin{table*}[tbp]
		\centering
		%	First row contains adversarial success rate for statistical-based method. 
		\caption{Imitation-based attack results based on 1000 generated curves. Results show the average ASR (in $[0,1]$) for start point and start sequence generation methods. Results presented are in the form of (mean $\pm$ standard deviation). The mean and standard deviation are calculated based on the aggregation strategy mentioned in Section \ref{imitateexp}.}
		\label{imitationres2}
		\begin{tabular}{cccccc}
			&          & \multicolumn{2}{c}{Balabit} & \multicolumn{2}{c}{TWOS}    \\ \hline
			Generation method                   & Sequence length   & SVM          & 1DCNN        & SVM          & 1DCNN        \\ \hline
			\multicolumn{2}{c}{Statistics-based baseline} & $0.6154 $      & $0.3183$       & $0.3895 $      &$ 0.264  $      \\ \hline
			\multirow{2}{*}{Start point}        & 50       & $0.41107\pm0.12024$ & $0.64551\pm0.11390$ & $0.38474 \pm0.078521$& $0.38836\pm0.030027$ \\
			& 100      &$ 0.541  \pm 0.11335 $    & $0.68878\pm0.073889$ & $0.39373\pm0.042965$ & $0.39726\pm0.043120$ \\\hline
			\multirow{2}{*}{Start sequence}    & 50       & $0.37581\pm0.117679$ & $0.55044\pm0.12358$ &$ 0.38193\pm0.074415 $&$ 0.40935\pm0.021899$  \\
			& 100      & $0.41796\pm0.12027$ & $0.69036\pm0.089278$ &$ 0.38628 \pm0.042309$&$0.40211\pm0.042972$ \\ \hline
		\end{tabular}
	\end{table*}
	
	%\begin{table*}[tbp]
	%	\centering
	%	%	First row contains adversarial success rate for statistical-based method. 
	%	\caption{Imitation-based attack results based on 1000 generated curves. Results show the average ASR (in $[0,1]$) for start point and start sequence generation methods. Results presented are in the form of (mean $\pm$ standard deviation) across the aggregated settings mentioned in Section \ref{imitateexp}.}
	%	\label{imitationres2}
	%	\begin{tabular}{cccccc}
	%		&          & \multicolumn{2}{c}{Balabit} & \multicolumn{2}{c}{TWOS}    \\ \hline
	%		Generation method                   & Sequence length   & SVM          & 1DCNN        & SVM          & 1DCNN        \\ \hline
	%		\multicolumn{2}{c}{Statistics-based baseline} & $0.6154 $      & $0.3183$       & $0.3895 $      &$ 0.264  $      \\ \hline
	%		\multirow{2}{*}{Start point}        & 50       & $0.41107$ & $0.64551$ & $0.38474$& $0.38836$ \\
	%		& 100      &$ 0.541 $    & $0.68878$ & $0.39373$ & $0.39726$ \\\hline
	%		\multirow{2}{*}{Start sequence}    & 50       & $0.37581$ & $0.55044$ &$ 0.38193 $&$ 0.40935$  \\
	%		& 100      & $0.41796$ & $0.69036$ &$ 0.38628 $&$0.40211$ \\ \hline
	%	\end{tabular}
	%\end{table*}
	
	For both the Balabit and TWOS datasets, the neural network-based attacks were consistently better than the statistics-based attacks against a 1DCNN target classifier. For the Balabit dataset, neural network-based methods were around $24\%$ to $38\%$ better than the baseline. For the TWOS dataset, neural network-based methods were around $12\%$ to $14\%$ better than the baseline. For SVM with the TWOS dataset, the results were comparable, however for SVM with the Balabit dataset, the results were poorer. 
	
	ASR is generally higher for the Balabit dataset than the TWOS dataset. This could be because the Balabit dataset is less complex than the TWOS dataset, as mentioned in Section \ref{datasets}. During the data preprocessing phase, we encountered many more anomalies in the TWOS data, e.g., having more instances of repeated time and mouse coordinates, compared to the Balabit data. As such, the generator would be able to learn sequences more effectively with the Balabit dataset than the TWOS dataset.

	An interesting observation is that training a generator and generating sequences based on length 100 yields a higher ASR in general than a sequence length of 50. This indicates that discriminative traits of users are more pronounced on longer time scales. The experimental results show that even with a straightforward way of training a generator and generating samples after, imitation-based methods can still bypass the target classifiers to a reasonable extent and are able to impact the robustness of these models adversely.
	
	\begin{figure}[tbp]
		% Boxplot illustrating the variability of ASR. For the ``DV'' settings, we used sequences of seqlen 50. For the ``VEL'' settings, we used sequences of seqlen 100. For both cases, we used the ``Cluster'' regularization strategy and we generated the adversarial samples using the ``Start Point'' strategy.
		\centering
		\includegraphics[trim={0 0 0.4cm 0.1cm},clip,width=0.45\textwidth]{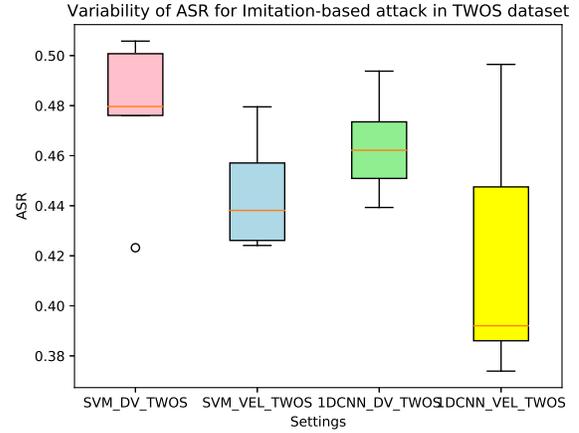}
		\caption{Boxplot illustrating the ASR variability. For the ``DV'' and ``VEL'' settings, we used sequences of seqlen 50 and 100 respectively. For both cases, we used the cluster regularization and we generated the adversarial samples using the start point strategy.}
		\label{imitationvariability}
	\end{figure}
	
	We analyse the stability of the imitation-based attack method, by repeating the training and generation procedures five times per setting, selecting two settings from the highest two ASR instances, to perform the variability test. We used the TWOS dataset for this set of experiments. The boxplot in Figure \ref{imitationvariability} summarizes our results. The setting when ``VEL'' was used yields a higher variability compared to the case of when ``DV'' is used, as can be seen in the figure where the former has a larger interquartile range (IQR), denoted by the coloured regions. Although the ``DV'' setting show a lower variability, it still obtained an IQR of approximately $2\%$. Hence, the imitation-based attack still has much room for improvement. Having said that, it still suffice to show that the target classifiers are not very robust to imitation-based attacks.

	\subsubsection{Surrogate-based Attack}
	%Note that we can only perform the perturbations only after the surrogate model has been trained.
	
	% In order to evaluate the performance of surrogate-based fooling, 
	%We used any other mouse trajectory that does not belong to the victim user to proceed with the perturbation.  
	%We used mouse trajectories that do not belong to the target user to proceed with the perturbation.
	Given the assumptions made regarding the threat model (discussed in Section \ref{threatmodel}), we only have mouse sequences from the targeted user. As such, we can only train our surrogate with positive samples from the target user and negative samples from other sources of data. To simulate this scenario, when performing experiments in the context of the Balabit dataset, we use the TWOS dataset sequences as the negative samples and vice versa. We followed the same approach for selecting a mouse sequence for perturbation. We conducted our experiments for this approach using velocity sequences.
	
	%\todo{Mention what dataset was used to train the surrogate}
	
	%We experimented with 2 different surrogate model architectures. One was a 3 layered GRU-RNN with 2 fully-connected (FC) layers at the end, with a Rectified Linear Unit (ReLU) activation after the first FC layer. The second uses only 3 FC layers placed one after another, with an ELU activation after the second layer. The first layer takes in a flattened sequence, where every \textit{odd} numbered node takes in the x-axis while the \textit{even} numbered node takes in the y-axis of the mouse sequence. The output of both architectures are 2-dimensional logits, each symbolizing the score for each decision (0 or 1). The outputs will be passed to a \textit{Cross-Entropy loss function} for loss calculation and eventual backpropagation.
	
	%The GRU-RNN model has a hidden dimension of 100, and a fully-connected layer at the end to convert the GRU-RNN outputs to 2-dimensional vectors. The FC surrogate model has hidden dimensions that are halved from the input dimensions at each FC layer except the last, where the output dimensions would be 2.
	
	The hyperparameters we used to train both of our surrogates are as follows. We used an Adam optimizer with a learning rate of 0.0005. We adopted a learning rate decay strategy every 10 epochs by a factor of 0.5, and the models were trained for 60 epochs. For the perturbation algorithm, we used an $\epsilon$ value of 0.001 in Equation \ref{fgsm}.
	%Table \ref{surrogateres} summarizes our results for the surrogate-based fooling method.
	\begin{table}[tbp]
		\centering
		%	First row contains fooling performance for statistical-based method.
		\caption{Surrogate-based attack results. Metrics reported are the ASR (in [0, 1]) when adversarial samples constructed based on surrogate model architectures are evaluated on the target classifiers.}
		\label{surrogateres}
		\begin{tabular}{ccccc}
			& \multicolumn{2}{c}{Balabit} & \multicolumn{2}{c}{TWOS}    \\ \cline{1-5} 
			Surrogate Variant & SVM          & 1DCNN        & SVM          & 1DCNN        \\ \hline
			Statistics-based & 0.6154       & 0.3183       & 0.3895       & 0.264        \\
			GRU-RNN surrogate & 0.6928       & 0.406        & 0.35265 & 0.36165 \\
			FC surrogate      & 0.6993       & 0.341       & 0.33996 & 0.43304 \\ \hline
		\end{tabular}
	\end{table}
	%As illustrated in Table \ref{surrogateres}, both the GRU-RNN and FC surrogate variants were able to fool Balabit target classifiers and TWOS 1DCNN target classifiers better than the statistical-based approach as shown in Table \ref{surrogateres}. Interestingly, the fooling performance of the GRU-RNN variant does not deviate too far from the FC variant, even though their architectures are vastly different. This shows that the attacker can use any arbitrary architecture as a surrogate model. 
	
	As illustrated in Table \ref{surrogateres}, both the GRU-RNN and FC surrogate variants were able to bypass the target classifiers better than the statistics-based approach, with the exception of the SVM target classifier using the TWOS dataset. Interestingly, the ASR of the GRU-RNN does not deviate very far from the FC variant, although their architectures are vastly different. This shows that the attacker can use any arbitrary surrogate model architecture, which will impact the robustness of the target classifiers adversely. 
	
	To show that the constructed sequences based on the surrogate models are the best that we can obtain, Table \ref{surrogateres2} shows the ASR of these constructed adversarial samples when evaluated against the surrogate models themselves. 
	\begin{table}[tbp]
		\centering
		%	First row contains fooling performance for statistical-based method.
		\caption{Surrogate-based attack results when the constructed adversarial samples were evaluated against the surrogate models. Metrics reported are the ASR, averaged across the users in the Balabit and TWOS datasets (in [0, 1]).}
		\label{surrogateres2}
		\begin{tabular}{ccccc}
			& \multicolumn{2}{c}{Balabit} & \multicolumn{2}{c}{TWOS}    \\ \cline{1-5} 
			Surrogate Variant & SVM          & 1DCNN        & SVM          & 1DCNN        \\ \hline		
			GRU-RNN surrogate & 0.995       & 0.995        & 0.849 & 0.848 \\
			FC surrogate      & 1.00       & 1.00       & 0.997 & 0.997 \\ \hline
		\end{tabular}
	\end{table}
	The table shows that high ASR can be achieved when white-box attacks are performed on the surrogate models. Recall that in Section \ref{method:surrogate}, we mentioned that we continue to perturb our input mouse sequence based on the loss of the surrogate model, calculated with respect to the input sequences. Hence, a high ASR for the surrogate models implies that our loss will be minimal and any further perturbation applied would be insignificant. As such, we have reached an ASR saturation point with the surrogate-based attacks in the given setting which was described in Section \ref{threatmodel}.
	
	% As the quality of the constructed adversarial samples are highly dependent on the ASR obtainable when tested on the surrogates
	
	In contrast to the imitation-based method, the weak upper limit does not apply here. This is because the surrogate-based method does not aim to imitate sequences but tries to find the region of space which is classified by the surrogate model as legitimate. The errors in the surrogate-based approach can be due to two factors. Firstly, in case of a mismatch between the target and surrogate architecture, intermediate feature representation differs. Thus, estimation of correctly classified regions will be unreliable, as seen in Table \ref{surrogateres}. Secondly, in the case of matching architectures, the errors are the result of the variability of decision boundaries associated with the different datasets used in the surrogate models and target classifiers. 
	%, once for training of the surrogate by the attacker, and then for tr. 
	In cases of non-convex optimization, as with neural networks, different weight initialization also matters. %instead find a set of perturbations that can be performed on any mouse sequence, so as to bypass the authenticator. Hence, this method makes use of the decision boundary of the surrogate models to calculate these perturbations, which is not restricted to the input space where the legitimate samples reside in. 

	\section{Discussion}
	\label{discuss}
	In this section, we discuss the relative efficiency of the explored approaches to perform adversarial attacks in the given setting. Next, we discuss in greater detail our approach to performing the Wilcoxon signed-ranked test and how we arrived at our conclusion for the test, which was mentioned in Section \ref{imitateexp}. Finally, we share our insights to an extension of our surrogate-based attack approach, in the event when the attacker knows the architecture of the victim's authentication model and a potential defensive mechanism to detect such adversarial attacks in this scenario.
	
	\subsection{Attack Strategy Selection}
	
	\label{ssec:strategyselection}
	%In Table \ref{imitationres2} it can be seen that with the TWOS dataset, which is a more challenging dataset, all imitation-based attacks perform roughly the same, with an intermediate ASR. 
	%The results from the imitation-based attacks are notably worse than those which could be obtained with access to the true authenticator (in contrast to our assumed threat model) and perform a white or black-box attack on it. 
	%This can be observed by measuring the attack performance on the surrogate from sequences obtained from the same surrogate itself.
	%This shows the challenges inherent in modelling the mouse movement sequences of a particular user.
	In Table \ref{imitationres2} it can be seen that with the TWOS dataset, which is a more challenging dataset, all imitation-based attacks perform roughly the same, approximately $38\%$ to $40\%$ ASR. 
	%The results from the imitation-based attacks are notably worse than those which could be obtained with access to the true authenticator (in contrast to our assumed threat model) and perform a white or black-box attack on it. 
	The results from the imitation-based attacks are notably worse than when the attacker could obtain access to the true authentication model (in contrast to our assumed threat model) and perform a white or black-box attack on it.
	%This can be observed by measuring the attack performance on the surrogate from sequences obtained from itself. 
	This shows the challenges inherent in modelling the mouse movement sequences of a particular user. However, a comparison of Tables \ref{imitationres2} and \ref{surrogateres} shows that most of the time using an imitation-based approach is better than training a surrogate which mismatches the architecture of the victim's authentication model. It should be noted that the surrogates in Table \ref{surrogateres} were deliberately chosen to be different from the architecture of the actual authentication model. 
	We refer the reader to Section \ref{ssec:surrogateextension} for more insights in this area.
	
	%We also note that the results regarding the Balabit dataset are more heterogeneous: the SVM-based authenticator can be well tricked by any of the surrogate-based attacks, whereas for the 1DCNN-based authenticator using a generative method is still the better option.
	
	For the statistics-based attack, it is consistently being outperformed by the other neural network-based attacks (imitation or surrogate-based), regardless of the dataset or the type of authentication model used. This shows that modelling the user's mouse movement sequences is even more challenging through a statistics-based approach. For SVM with the Balabit dataset, the surrogate-based attack is better while for the rest, the imitation-based is better. As such, using statistics to model a user's mouse movement sequences should be avoided.
	
	To summarise, both the imitation-based and surrogate-based approaches are viable options to perform adversarial attacks on the authentication model. This is because there are instances where when one approach fails, the other succeeds. Hence, a neural network-based attack is still a better option than a statistics-based one.
	
	%Hence, based on our results, we deduced that the surrogate-based attack is the most effective as it is more generalizable.

	\subsection{Does Representation or Regularization Matter?}
	\label{wilcoxon}
	
	% We investigated whether the representation of inputs or the chosen regularizarion strategy has a statistically measurable impact on the ASR.
	We investigated whether the input representation or regularization strategy has a statistically measurable impact on the ASR. 
	The Wilcoxon signed-ranked test was performed by forming pairs between experimental results that have the same settings, except for the variable being compared.\footnote{For example, if we would like to compare the effects of having \textit{differing sequence lengths}, we would form pairs between the results in which each of the values in the pair would come from length $50$ and length $100$, with all other settings being consistent.}
	
	At $5\%$ significance level, the critical z-value for a two-tailed test is $1.96$. During the comparison between the different model strategies, we compared between ABS and DV, ABS and VEL, and DV and VEL. During each of the tests, there were $48$ pairs present; for brevity, the calculated z-values were $0.45642$, $1.3282$, and $1.1641$ respectively. Thus, we concluded that 
	%we could not reject the null hypothesis as there is no significant evidence to do so. Therefore, this signifies that there is \textit{no significant impact by altering the input representation of the generator}.
	there is \textit{no significant evidence to claim that there is a significant impact by altering the input representation of the generator}.
	%, where it states that the difference between the pairs of values follows a symmetric distribution around zero
	
	In our comparison of the regularizing strategies, we compared between ``No'' and ``Cluster'', ``No'' and ``Derivative'', and ``Cluster'' and ``Derivative''. During each of the tests, there were $48$ pairs present; again, for brevity, the calculated z-values were $0.35385$, $0.087181$, and $0.29231$ respectively. Thus, we concluded that 
	%there is no significant evidence to reject the null hypothesis and therefore, there is \textit{no significant impact by altering the regularizing strategy}.
	there is \textit{no significant evidence to claim that there is significant impact by altering the regularizing strategy}.

	\subsection{Surrogate-based Attack Extension}
	\label{ssec:surrogateextension}
	
	%As an extension of the surrogate-based idea, 
	
	We also asked ourselves what happens if the attacker has knowledge of the architecture of the target classifiers and also the dataset that consists of the same group of users that were used to train them (e.g., if multiple users are using the victim's machine), but when one still does not have access to the parameters learned by the classifier.
	%Note that this dataset does not need to come from the training set of the target classifiers.
	%, perhaps through sniffing the other users' data as well
	Interestingly, if the attacker trains a surrogate with the same architecture as the target classifier, the ASR is much higher than the other two approaches, as evident in Table \ref{surrogateextension}. In the Balabit dataset, the ASR can reach $92.1\%$ for the SVM and $81.7\%$ for the 1DCNN. For the TWOS dataset, the ASR can reach $66.9\%$ for the SVM and $58.7\%$ for the 1DCNN. This is the highest result achieved among the proposed approaches.
	% even when comparing against the imitation-based approach. 
	%Hence, this shows that if the attacker knows the architecture of the target classifier and has the appropriate data, then the adversarial results are much higher compared to the case of mismatched architectures. 
	Therefore, having any form of query access to the target classifier is not required to achieve these relatively high ASR results, although if the attacker fails to obtain any of this information, it is clear that the ASR will suffer. Yet the results are not close to 100\%, showing that performance suffers when one cannot query the target classifier as in \cite{Papernot2016}.
	% (potentially worse than the statistics-based approach).
	\begin{table}[tbp]
		\centering
		% 	Surrogate-based attack results when the attacker has access to the target classifier's model architecture and the user classes that the authenticator model trained on. The bolded values highlights the high ASR when a surrogate of the same architecture is used.
		\caption{Surrogate-based attack results when the attacker has access to the target classifier's model architecture. The bolded values highlights the high ASR obtained when a surrogate model of the same architecture as the target classifier is used.}
		\label{surrogateextension}
		\begin{tabular}{ccccc}
			& \multicolumn{2}{c}{Balabit} & \multicolumn{2}{c}{TWOS}    \\ \cline{1-5} 
			Surrogate Variant & SVM          & 1DCNN        & SVM          & 1DCNN        \\ \hline
			Statistics-based & 0.6154       & 0.3183       & 0.3895       & 0.264        \\
			SVM surrogate 	& \textbf{0.92108}       & 0.39801        & \textbf{0.66970} 	& 0.22278 \\
			1DCNN surrogate      & 0.60583       & \textbf{0.81707}       & 0.36881 & \textbf{0.58710} \\ \hline
		\end{tabular}
	\end{table}
	
	The results presented in Table \ref{surrogateextension} suggest a possible detection strategy against surrogate-based attacks using a single surrogate for our threat model. Namely, one employs a probabilistic average of multiple authentication models, where at every timestep one randomly decides which authentication model to use. The difference in ASR would be reflected in the alert frequency which would change when the models are switched (see Table \ref{alerts}). Increasing alert rates (compared to using legitimate user's data) due to a shift in test distribution (covariate shifts) would affect all models to some degree, but an adversarial attack by a surrogate would affect one model less than all of the others. Hence, detection can be performed by checking the alert rates across the models to observe whether one model has an exceptionally low alert rate, while the others have a much higher alert rate.
	% Table \ref{alerts} illustrates this point.

	\begin{table}[tb]
		\caption{Alert rates (in [0, 1]) obtained for corresponding experimental settings. Covariate shifts in the data are made by performing affine rotations ranging from 45\textdegree ~to 90\textdegree.}
		\label{alerts}
		\centering
		\begin{tabular}{ccccc}
			& \multicolumn{2}{c}{Balabit} & \multicolumn{2}{c}{TWOS}    \\ \hline
			Experiment Setting           & SVM          & 1DCNN        & SVM          & 1DCNN        \\ \hline
			Legitimate user's data       & 0.412     & 0.302     & 0.301      & 0.477     \\
			Surrogate-based attack (SVM) & 0.0774       & 0.660       & 0.375 & 0.803 \\
			Covariate shift              & 0.444     & 0.352     & 0.525     & 0.561     \\ \hline
		\end{tabular}
	\end{table}
	
	\section{Conclusion}
	\label{conclude}
	%We have shown that behaviour-based mouse dynamics authenticators are not very robust to adversarial attacks, when tested in a realistic setting (see Section \ref{threatmodel}). We proposed different strategies that can be used by a potential attacker to launch synthetically-generated adversarial samples, either through imitation-based, surrogate-based or even statistics-based approaches. Although the generation of mouse sequences is a difficult task, and the effectiveness of adversarial attacks can be improved, it is sufficient to show that the robustness of these authenticators can be adversely affected. 
	% We also show that these adversarial attacks can be generalized across different contexts, from our experimentation with the Balabit and TWOS datasets. 
	%To the best of our knowledge, the concept of adversarial attacks has not yet been studied in this setting, so we also provide a baseline in attacking the robustness of such authenticators. 
	
	In conclusion, we proposed different strategies that a potential attacker can use to launch synthetically-generated adversarial samples, either through imitation-based, surrogate-based or statistics-based approaches. Based on the experimental results, we show that neural network-based attacks (imitation or surrogate based) are better performing than a statistics-based attacks. Although the generation of mouse sequences is a difficult task, and the proposed adversarial attacks have their flaws, it is sufficient to show that the robustness of these authentication models can be adversely affected even when tested in a realistic setting (see Section \ref{threatmodel}). 
	Also, we have shown that if the attacker could guess the architecture of the authentication model correctly, its robustness would be greatly affected even without any form of access to it.
	One can also infer from our results, particularly those presented in Tables \ref{imitationres2}, \ref{surrogateres}, and \ref{surrogateextension}, that in a realistic setting in which the authentication model is inaccessible, attacking a system based on behavioural features is harder than, for example, copying a fingerprint. 

	\section*{Acknowledgment}
	
	This work was supported by both ST Electronics and the National Research Foundation (NRF), Prime Minister’s Office, Singapore under Corporate Laboratory @ University Scheme (Programme Title: STEE Infosec-SUTD Corporate Laboratory). Alexander Binder also gratefully acknowledges the support by PIE-SGP-AI-2018-01.
	
	%\section*{References}
	
	%Please number citations consecutively within brackets \cite{b1}. The 
	%sentence punctuation follows the bracket \cite{b2}. Refer simply to the reference 
	%number, as in \cite{b3}---do not use ``Ref. \cite{b3}'' or ``reference \cite{b3}'' except at 
	%the beginning of a sentence: ``Reference \cite{b3} was the first $\ldots$''
	%
	%Number footnotes separately in superscripts. Place the actual footnote at 
	%the bottom of the column in which it was cited. Do not put footnotes in the 
	%abstract or reference list. Use letters for table footnotes.
	%
	%Unless there are six authors or more give all authors' names; do not use 
	%``et al.''. Papers that have not been published, even if they have been 
	%submitted for publication, should be cited as ``unpublished'' \cite{b4}. Papers 
	%that have been accepted for publication should be cited as ``in press'' \cite{b5}. 
	%Capitalize only the first word in a paper title, except for proper nouns and 
	%element symbols.
	%
	%For papers published in translation journals, please give the English 
	%citation first, followed by the original foreign-language citation \cite{b6}.
	
	\bibliographystyle{IEEEtran}
	\bibliography{references,nonmendeleyref}

\begin{thebibliography}{10}
\providecommand{\url}[1]{#1}
\csname url@samestyle\endcsname
\providecommand{\newblock}{\relax}
\providecommand{\bibinfo}[2]{#2}
\providecommand{\BIBentrySTDinterwordspacing}{\spaceskip=0pt\relax}
\providecommand{\BIBentryALTinterwordstretchfactor}{4}
\providecommand{\BIBentryALTinterwordspacing}{\spaceskip=\fontdimen2\font plus
\BIBentryALTinterwordstretchfactor\fontdimen3\font minus
  \fontdimen4\font\relax}
\providecommand{\BIBforeignlanguage}[2]{{%
\expandafter\ifx\csname l@#1\endcsname\relax
\typeout{** WARNING: IEEEtran.bst: No hyphenation pattern has been}%
\typeout{** loaded for the language `#1'. Using the pattern for}%
\typeout{** the default language instead.}%
\else
\language=\csname l@#1\endcsname
\fi
#2}}
\providecommand{\BIBdecl}{\relax}
\BIBdecl

\bibitem{bhattacharyya2009biometric}
D.~Bhattacharyya, R.~Ranjan, F.~Alisherov, M.~Choi \emph{et~al.}, ``Biometric
  authentication: A review,'' \emph{International Journal of u-and e-Service,
  Science and Technology}, vol.~2, no.~3, pp. 13--28, 2009.

\bibitem{monrose2000keystroke}
F.~Monrose and A.~D. Rubin, ``Keystroke dynamics as a biometric for
  authentication,'' \emph{Future Generation computer systems}, vol.~16, no.~4,
  pp. 351--359, 2000.

\bibitem{Ahmed2007}
\BIBentryALTinterwordspacing
A.~A.~E. Ahmed and I.~Traore, ``{A New Biometric Technology Based on Mouse
  Dynamics},'' \emph{IEEE Transactions on Dependable and Secure Computing},
  vol.~4, no.~3, pp. 165--179, 2007. [Online]. Available:
  \url{http://ieeexplore.ieee.org/document/4288179/}
\BIBentrySTDinterwordspacing

\bibitem{Chong2018}
\BIBentryALTinterwordspacing
P.~Chong, Y.~X.~M. Tan, J.~Guarnizo, Y.~Elovici, and A.~Binder, ``{Mouse
  Authentication Without the Temporal Aspect – What Does a 2D-CNN Learn?}''
  \emph{2018 IEEE Security and Privacy Workshops (SPW)}, pp. 15--21, 2018.
  [Online]. Available: \url{https://ieeexplore.ieee.org/document/8424627/}
\BIBentrySTDinterwordspacing

\bibitem{Feher2012}
\BIBentryALTinterwordspacing
C.~Feher, Y.~Elovici, R.~Moskovitch, L.~Rokach, and A.~Schclar, ``{User
  identity verification via mouse dynamics},'' \emph{Information Sciences},
  vol. 201, pp. 19--36, 2012. [Online]. Available:
  \url{http://dx.doi.org/10.1016/j.ins.2012.02.066}
\BIBentrySTDinterwordspacing

\bibitem{Gamboa2004}
H.~Gamboa and A.~Fred, ``{A behavioral biometric system based on human computer
  interaction},'' \emph{Proceedings of SPIE}, vol. 5404, no.~i, pp. 381--392,
  2004.

\bibitem{Kasprowski2018}
P.~Kasprowski and K.~Harezlak, ``{Fusion of eye movement and mouse dynamics for
  reliable behavioral biometrics},'' \emph{Pattern Analysis and Applications},
  vol.~21, no.~1, pp. 91--103, 2018.

\bibitem{Mondal2017}
\BIBentryALTinterwordspacing
S.~Mondal and P.~Bours, ``{A study on continuous authentication using a
  combination of keystroke and mouse biometrics},'' \emph{Neurocomputing}, vol.
  230, no. November 2016, pp. 1--22, 2017. [Online]. Available:
  \url{http://dx.doi.org/10.1016/j.neucom.2016.11.031}
\BIBentrySTDinterwordspacing

\bibitem{Shen2012}
C.~Shen, Z.~Cai, and X.~Guan, ``{Continuous authentication for mouse dynamics:
  A pattern-growth approach},'' \emph{Proceedings of the International
  Conference on Dependable Systems and Networks}, 2012.

\bibitem{Shen2013}
C.~Shen, Z.~Cai, X.~Guan, Y.~Du, and R.~A. Maxion, ``{User authentication
  through mouse dynamics},'' \emph{IEEE Transactions on Information Forensics
  and Security}, vol.~8, no.~1, pp. 16--30, 2013.

\bibitem{Sayed2013}
B.~Sayed, I.~Traore, I.~Woungang, and M.~S. Obaidat, ``{Biometric
  authentication using mouse gesture dynamics},'' \emph{IEEE Systems Journal},
  vol.~7, no.~2, pp. 262--274, 2013.

\bibitem{Mo2018}
\BIBentryALTinterwordspacing
F.~Mo, S.~Xiong, S.~Yi, Q.~Yi, and A.~Zhang, \emph{{Intelligent Computing and
  Internet of Things}}.\hskip 1em plus 0.5em minus 0.4em\relax Springer
  Singapore, 2018, vol. 924. [Online]. Available:
  \url{http://link.springer.com/10.1007/978-981-13-2384-3}
\BIBentrySTDinterwordspacing

\bibitem{5291887}
Y.~Aksari and H.~Artuner, ``Active authentication by mouse movements,'' in
  \emph{2009 24th International Symposium on Computer and Information
  Sciences}, Sept 2009, pp. 571--574.

\bibitem{5337543}
P.~Bours and C.~J. Fullu, ``A login system using mouse dynamics,'' in
  \emph{2009 Fifth International Conference on Intelligent Information Hiding
  and Multimedia Signal Processing}, Sept 2009, pp. 1072--1077.

\bibitem{Zheng2011}
\BIBentryALTinterwordspacing
N.~Zheng, A.~Paloski, and H.~Wang, ``{An efficient user verification system via
  mouse movements},'' \emph{Proceedings of the 18th ACM conference on Computer
  and communications security - CCS '11}, no. February, p. 139, 2011. [Online].
  Available: \url{http://dl.acm.org/citation.cfm?doid=2046707.2046725}
\BIBentrySTDinterwordspacing

\bibitem{Papernot2016}
\BIBentryALTinterwordspacing
N.~Papernot, P.~McDaniel, I.~Goodfellow, S.~Jha, Z.~B. Celik, and A.~Swami,
  ``Practical black-box attacks against machine learning,'' in
  \emph{Proceedings of the 2017 ACM on Asia Conference on Computer and
  Communications Security}, ser. ASIA CCS '17.\hskip 1em plus 0.5em minus
  0.4em\relax New York, NY, USA: ACM, 2017, pp. 506--519. [Online]. Available:
  \url{http://doi.acm.org/10.1145/3052973.3053009}
\BIBentrySTDinterwordspacing

\bibitem{Papernot2016a}
\BIBentryALTinterwordspacing
N.~Papernot, P.~McDaniel, S.~Jha, M.~Fredrikson, Z.~Celik, and A.~Swami, ``The
  limitations of deep learning in adversarial settings,'' in \emph{2016 IEEE
  European Symposium on Security and Privacy (EuroS\&P)}.\hskip 1em plus 0.5em
  minus 0.4em\relax Los Alamitos, CA, USA: IEEE Computer Society, mar 2016, pp.
  372--387. [Online]. Available:
  \url{https://doi.ieeecomputersociety.org/10.1109/EuroSP.2016.36}
\BIBentrySTDinterwordspacing

\bibitem{Goodfellow2014}
\BIBentryALTinterwordspacing
I.~J. Goodfellow, J.~Shlens, and C.~Szegedy, ``{Explaining and Harnessing
  Adversarial Examples},'' pp. 1--11, 2014. [Online]. Available:
  \url{http://arxiv.org/abs/1412.6572}
\BIBentrySTDinterwordspacing

\bibitem{carlini2017towards}
N.~Carlini and D.~Wagner, ``Towards evaluating the robustness of neural
  networks,'' in \emph{2017 IEEE Symposium on Security and Privacy (SP)}.\hskip
  1em plus 0.5em minus 0.4em\relax IEEE, 2017, pp. 39--57.

\bibitem{brendel2017decision}
W.~Brendel, J.~Rauber, and M.~Bethge, ``Decision-based adversarial attacks:
  Reliable attacks against black-box machine learning models,'' \emph{arXiv
  preprint arXiv:1712.04248}, 2017.

\bibitem{evtimov2017robust}
I.~Evtimov, K.~Eykholt, E.~Fernandes, T.~Kohno, B.~Li, A.~Prakash, A.~Rahmati,
  and D.~Song, ``Robust physical-world attacks on machine learning models,''
  \emph{arXiv preprint arXiv:1707.08945}, 2017.

\bibitem{szegedy2013intriguing}
C.~Szegedy, W.~Zaremba, I.~Sutskever, J.~Bruna, D.~Erhan, I.~Goodfellow, and
  R.~Fergus, ``Intriguing properties of neural networks,'' \emph{arXiv preprint
  arXiv:1312.6199}, 2013.

\bibitem{balabit}
A.~F\"ul\"op, L.~Kov\'acs, T.~Kurics, and E.~Windhager-Pokol, ``Balabit mouse
  dynamics challenge data set,''
  \url{https://github.com/balabit/Mouse-Dynamics-Challenge}, 2016.

\bibitem{binder2016layer}
A.~Binder, G.~Montavon, S.~Lapuschkin, K.-R. M{\"u}ller, and W.~Samek,
  ``Layer-wise relevance propagation for neural networks with local
  renormalization layers,'' in \emph{International Conference on Artificial
  Neural Networks}.\hskip 1em plus 0.5em minus 0.4em\relax Springer, 2016, pp.
  63--71.

\bibitem{bach2015pixel}
S.~Bach, A.~Binder, G.~Montavon, F.~Klauschen, K.-R. M{\"u}ller, and W.~Samek,
  ``On pixel-wise explanations for non-linear classifier decisions by
  layer-wise relevance propagation,'' \emph{PloS one}, vol.~10, no.~7, p.
  e0130140, 2015.

\bibitem{Rosenberg2018}
I.~Rosenberg, A.~Shabtai, L.~Rokach, and Y.~Elovici, ``Generic black-box
  end-to-end attack against state of the art api call based malware
  classifiers,'' in \emph{International Symposium on Research in Attacks,
  Intrusions, and Defenses}.\hskip 1em plus 0.5em minus 0.4em\relax Springer,
  2018, pp. 490--510.

\bibitem{homoliak2018}
I.~Homoliak, F.~Toffalini, J.~Guarnizo, Y.~Elovici, and M.~Ochoa, ``Insight
  into insiders and it: A survey of insider threat taxonomies, analysis,
  modeling, and countermeasures,'' 11 2018.

\bibitem{Williams1989}
\BIBentryALTinterwordspacing
R.~J. Williams and D.~Zipser, ``{A Learning Algorithm for Continually Running
  Fully Recurrent Neural Networks},'' \emph{Neural Computation}, vol.~1, no.~2,
  pp. 270--280, 1989. [Online]. Available:
  \url{http://www.mitpressjournals.org/doi/10.1162/neco.1989.1.2.270}
\BIBentrySTDinterwordspacing

\bibitem{paszke2017automatic}
A.~Paszke, S.~Gross, S.~Chintala, G.~Chanan, E.~Yang, Z.~DeVito, Z.~Lin,
  A.~Desmaison, L.~Antiga, and A.~Lerer, ``Automatic differentiation in
  pytorch,'' in \emph{NIPS-W}, 2017.

\bibitem{article}
A.~Harilal, F.~Toffalini, I.~Homoliak, J.~Castellanos, J.~Guarnizo, S.~Mondal,
  and M.~Ochoa, ``The wolf of sutd (twos): A dataset of malicious insider
  threat behavior based on a gamified competition,'' vol.~9, 03 2018.

\bibitem{scikitlearn}
F.~Pedregosa, G.~Varoquaux, A.~Gramfort, V.~Michel, B.~Thirion, O.~Grisel,
  M.~Blondel, P.~Prettenhofer, R.~Weiss, V.~Dubourg, J.~Vanderplas, A.~Passos,
  D.~Cournapeau, M.~Brucher, M.~Perrot, and E.~Duchesnay, ``Scikit-learn:
  Machine learning in {P}ython,'' \emph{Journal of Machine Learning Research},
  vol.~12, pp. 2825--2830, 2011.

\bibitem{wilcoxon1945individual}
F.~Wilcoxon, ``Individual comparisons by ranking methods,'' \emph{Biometrics
  bulletin}, vol.~1, no.~6, pp. 80--83, 1945.

\end{thebibliography}
	
	%\bibliography{nonmendeleyref}
	%\vspace{12pt}
	%\color{red}
	%IEEE conference templates contain guidance text for composing and formatting conference papers. Please ensure that all template text is removed from your conference paper prior to submission to the conference. Failure to remove the template text from your paper may result in your paper not being published.
	
\end{document}